\documentclass[conference]{IEEEtran}
\IEEEoverridecommandlockouts    

\newcommand{\isAnonymous}{0} 

\usepackage{multirow}
\usepackage{lipsum}
\usepackage{amsmath}
\usepackage{amssymb}
\usepackage[ruled,vlined]{algorithm2e}
\usepackage{graphicx}
\graphicspath{{./Figures/}}
\usepackage[margin=1in]{geometry}

\usepackage{amsmath}
\usepackage{amssymb}

\usepackage{booktabs}
\usepackage{siunitx}

\usepackage{float}

\usepackage{graphicx}
\graphicspath{{./Figures/}}

\usepackage{etoolbox} 
\usepackage{stackengine}

\newcommand{\maybeCensor}[5]{%
    \ifnum\isAnonymous=1%
        \stackinset{l}{#1}{b}{#2}{\rule{#3}{#4}}{#5}%
    \else%
        #5%
    \fi%
}

\usepackage{acronym}

\usepackage{tikz}
\usepackage{pgfgantt}      
\usepackage{pgfcalendar}   
\usepackage{xfp}           
\usetikzlibrary{calendar}        
\usetikzlibrary{arrows.meta}     
\usetikzlibrary{positioning}     
\usetikzlibrary{shapes}          
\usetikzlibrary{plotmarks}       

\usepackage{pgfplots}
\usepgfplotslibrary{statistics}
\usepgfplotslibrary{dateplot}    
\usepgfplotslibrary{groupplots}
\usepackage{pgfplotstable}
\usepgfplotslibrary{statistics}
\usepgfplotslibrary{fillbetween}

\pgfplotsset{
	compat=1.18,
	legend style={
			font=\fontsize{7}{11}\selectfont
		},
	label style={font=\fontsize{7}{11}\selectfont},
	standard-plot/.style={
			grid=major,
			grid style={dashdotted},
		},
	line-plot/.style={
			standard-plot,
			width=\columnwidth,
			height=0.5\columnwidth,
		},
	group-plot/.style={
			width=\columnwidth,
			height=0.495\columnwidth,
		},
	group-plot-3/.style={
			group-plot,
			group style={
					group size=1 by 3,
					xlabels at=edge bottom,
					xticklabels at=edge bottom,
					vertical sep=0cm,
				},
		},
	group-plot-4/.style={
			group-plot,
			group style={
					group size=1 by 4,
					xlabels at=edge bottom,
					xticklabels at=edge bottom,
					vertical sep=0cm,
				},
		},
	generic-linestyle/.style={thick},
	line-1/.style={generic-linestyle, TUMBlack},
	line-2/.style={generic-linestyle, TUMOrange},
	line-3/.style={generic-linestyle, TUMBlue},
	line-4/.style={generic-linestyle, TUMGreen},
	line-5/.style={line-1, dashed},
	line-6/.style={line-2, dashed},
	line-7/.style={line-3, dashed},
	line-8/.style={line-4, dashed},
}

\usepackage{hyperref}

\definecolor{TUMBlue}{RGB}{0,101,189}
\definecolor{TUMWhite}{RGB}{255,255,255}%
\definecolor{TUMBlack}{RGB}{0,0,0}%
\definecolor{TUMBlue1}{RGB}{0,51,89}
\definecolor{TUMBlue2}{RGB}{0,82,147}
\definecolor{TUMGray1}{RGB}{51,51,51}%
\definecolor{TUMGray2}{RGB}{127,127,127}%
\definecolor{TUMGray3}{RGB}{204,204,204}%
\definecolor{TUMBlue3}{RGB}{100,160,200}
\definecolor{TUMBlue4}{RGB}{152,198,234}
\definecolor{TUMIvory}{RGB}{218,215,203}%
\definecolor{TUMOrange}{RGB}{227,114,34}%
\definecolor{TUMGreen}{RGB}{162,173,0}%

\acrodef{mpc}[MPC]{Model Predictive Control}
\acrodef{hd}[HD]{Handling Diagram}
\acrodef{ehd}[EHD]{Empirical Handling Diagram}
\acrodef{msnn}[MS-NN]{Model-Structured Neural Network}
\acrodef{lstm}[LSTM]{Long Short-Term Memory}
\acrodef{shap}[SHAP]{SHapley Additive exPlanations}
\acrodef{rmse}[RMSE]{Root Mean Squared Error}
\acrodef{mae}[MAE]{Mean Absolute Error}
\acrodef{pinn}[PINN]{Physics-Informed Neural Network}
\acrodef{nn}[NN]{Neural Network}
\acrodef{fnn}[FNN]{Feedforward Neural Network}
\acrodef{cnn}[CNN]{Convolutional Neural Network}
\acrodef{fvu}[FVU]{Fraction of Variance Unexplained}
\acrodef{rslcpp}[RSLCPP]{ROS Simulation Library for C++}
\acrodef{a2rl}[A2RL]{Abu Dhabi Autonomous Racing League}
\acrodef{iac}[IAC]{Indy Autonomous Challenge}
\acrodef{tam}[TAM]{TUM Autonomous Motorsports}

\title{\LARGE \bf Benchmarking Empirical and Learning-Based Approaches for Feedforward Steering Control in Autonomous Racing}

\author{
	\ifnum\isAnonymous=0
		\parbox{\textwidth}{%
			\centering
			Georg Jank$^{1}$, Mattia Piccinini$^{2}$, Sebastian Wenk$^1$, Phillip Pitschi$^{1}$, Johannes Betz$^{2}$, Boris Lohmann$^1$%
		}%
		\thanks{$^{1}$Chair of Automatic Control, Department of Engineering Physics and Computation, Technical University of Munich,
			Boltzmannstraße 15, 85748 Garching bei München, Germany
			{\tt\small georg.jank@tum.de}
			}%
		\thanks{$^{2}$Professorship of Autonomous Vehicle Systems (AVS), Department of Mobility Systems Engineering, Technical University of Munich,
			Boltzmannstraße 15, 85748 Garching bei München, Germany
			}%
	\fi
}

\begin{document}
\bstctlcite{IEEEexample:BSTcontrol}


\maketitle
\thispagestyle{empty}
\pagestyle{empty}

\enlargethispage{-4\baselineskip}
\begin{tikzpicture}[remember picture, overlay]
  \node[anchor=south, yshift=5mm, inner sep=0pt] at (current page.south) {%
    \fbox{\parbox{\dimexpr\textwidth-2\fboxsep-2\fboxrule\relax}{%
      \small\copyright 2026 IEEE. Personal use of this material is permitted. Permission from IEEE must be obtained for all other uses, in any current or future media, including reprinting/republishing this material for advertising or promotional purposes, creating new collective works, for resale or redistribution to servers or lists, or reuse of any copyrighted component of this work in other works.%
    }}%
  };
\end{tikzpicture}

\begin{abstract}
Feedforward steering control is a key component of hierarchical control architectures for autonomous racing. The goal is to reduce steering corrections from the feedback controllers by predicting the vehicle's inverse lateral dynamics. This paper presents a systematic benchmark of two learning-based and two empirical (analytical) feedforward steering controllers. We introduce a new \acf{ehd} formulation based on a polynomial surface fit that captures velocity-dependent nonlinear steering behavior with minimal parametrization. 
We test the feedforward controllers in a high-fidelity simulation framework based on the real-world Abu Dhabi Autonomous Racing League competition, using a high-fidelity double-track vehicle dynamics simulator. 
Open-loop evaluation shows that the learning-based controllers achieve the lowest prediction errors; 
however, closed-loop testing reveals that this improved accuracy does not translate into superior path tracking performance or lap times, even after iterative fine-tuning. 
In contrast, the proposed EHD approach achieves the best overall closed-loop robustness and lap time, highlighting the necessity of evaluating feedforward strategies within the complete trajectory planning and control software stack.
Our code is available at \url{https://github.com/TUMRT/steering_ff_control}.
\end{abstract}
\section{Introduction}
	\label{sec:introduction}
\begin{figure}
        \centering
        \def\svgwidth{\columnwidth}
        \ifnum \isAnonymous=1
            \stackinset{l}{0.45\columnwidth}{b}{20pt}{\rule{1.1cm}{1.7cm}}{%
\begingroup%
  \makeatletter%
  \providecommand\color[2][]{%
    \errmessage{(Inkscape) Color is used for the text in Inkscape, but the package 'color.sty' is not loaded}%
    \renewcommand\color[2][]{}%
  }%
  \providecommand\transparent[1]{%
    \errmessage{(Inkscape) Transparency is used (non-zero) for the text in Inkscape, but the package 'transparent.sty' is not loaded}%
    \renewcommand\transparent[1]{}%
  }%
  \providecommand\rotatebox[2]{#2}%
  \newcommand*\fsize{\dimexpr\f@size pt\relax}%
  \newcommand*\lineheight[1]{\fontsize{\fsize}{#1\fsize}\selectfont}%
  \ifx\svgwidth\undefined%
    \setlength{\unitlength}{1201.8290362bp}%
    \ifx\svgscale\undefined%
      \relax%
    \else%
      \setlength{\unitlength}{\unitlength * \real{\svgscale}}%
    \fi%
  \else%
    \setlength{\unitlength}{\svgwidth}%
  \fi%
  \global\let\svgwidth\undefined%
  \global\let\svgscale\undefined%
  \makeatother%
  \begin{picture}(1,0.57587298)%
    \lineheight{1}%
    \setlength\tabcolsep{0pt}%
    \put(0,0){\includegraphics[width=\unitlength,page=1]{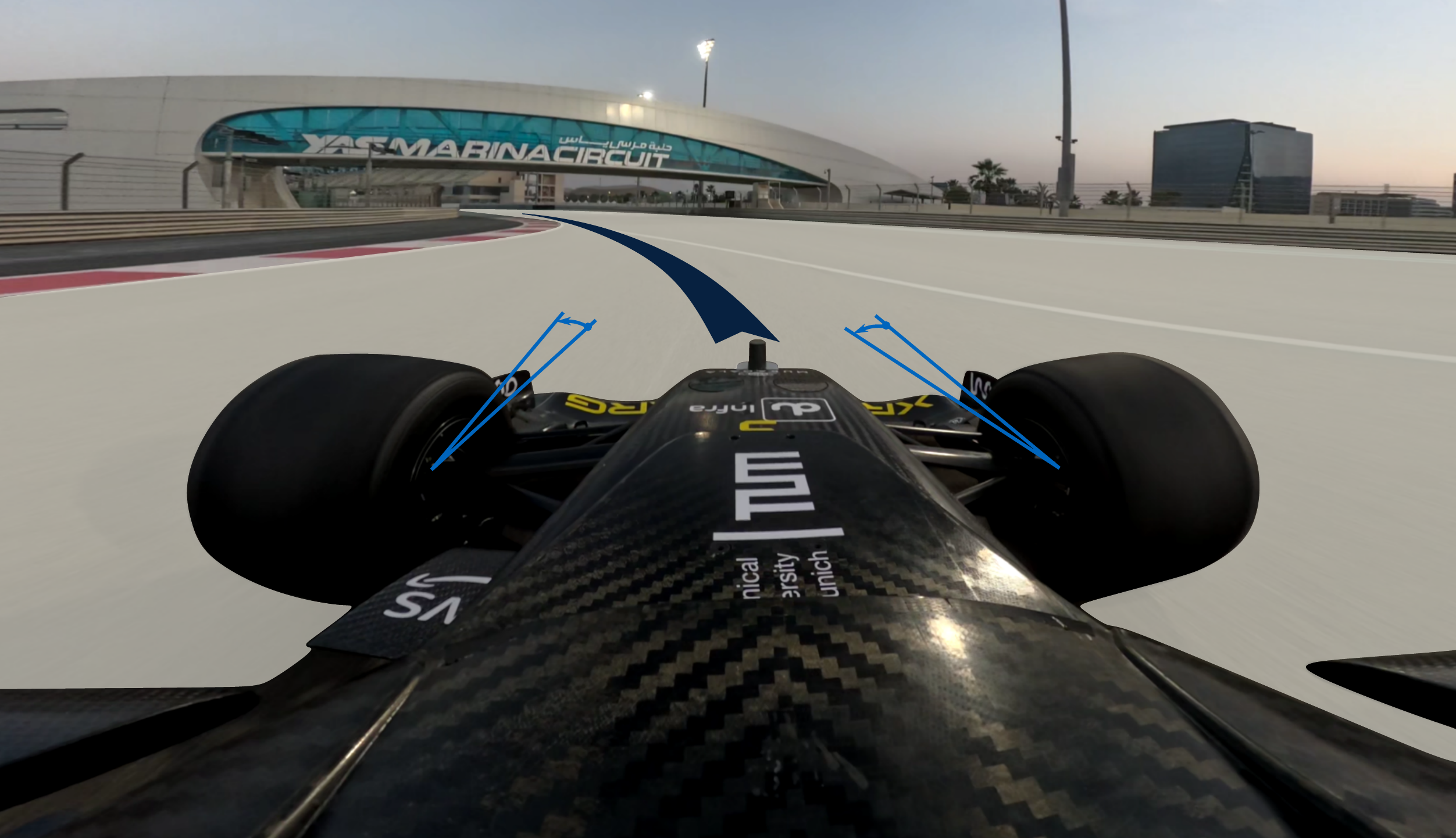}}%
    \put(0.64078335,0.35178276){\color[rgb]{0,0.39607843,0.74117647}\makebox(0,0)[lt]{\lineheight{1.25}\smash{\begin{tabular}[t]{l}$\delta_\mathrm{ff}=?$\end{tabular}}}}%
    \put(0.25688529,0.34859002){\color[rgb]{0,0.39607843,0.74117647}\makebox(0,0)[lt]{\lineheight{1.25}\smash{\begin{tabular}[t]{l}$\delta_\mathrm{ff}=?$\end{tabular}}}}%
    \put(0.5063024,0.38718416){\color[rgb]{0.02745098,0.12941176,0.25098039}\makebox(0,0)[lt]{\lineheight{1.25}\smash{\begin{tabular}[t]{l}Target\end{tabular}}}}%
  \end{picture}%
\endgroup%
            }
        \else
            
        \fi
        \caption{Onboard of the Dallara EAV24 at the Abu Dhabi circuit. We benchmark existing and new feedforward controllers that compute the steering command $\delta_{\mathrm{ff}}$ to track the desired trajectory provided by a high-level \ac{mpc}.}
        \label{fig:motivation}
    \end{figure}
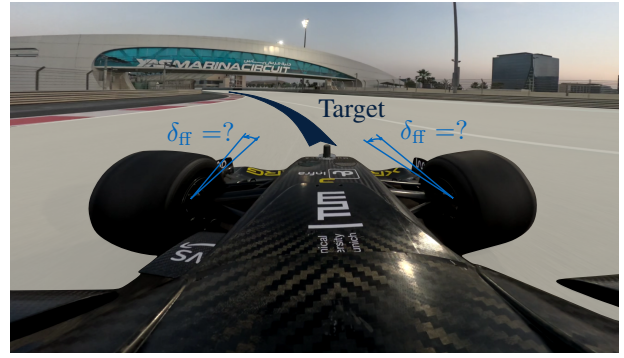

    Autonomous racing competitions, such as the \ac{a2rl}, provide a safe yet highly competitive environment for the development of vehicle control algorithms at the limits of tire-road friction~\cite{betz2022autonomous} (see Fig. \ref{fig:motivation}). A widely adopted strategy for autonomous vehicle control is to combine a high-level \ac{mpc} with low-level lateral and longitudinal controllers that track the \ac{mpc} path, velocity, and acceleration profiles \cite{Wischnewski2021,Spielberg2019,Piccinini2025ICRA,piccinini2023predictive,He2020}. Among the low-level lateral controllers, feedforward steering controllers aim to approximate the inverse lateral dynamics of the vehicle and minimize corrective feedback contributions. While previous work has evaluated learning-based feedforward steering controllers in open-loop operation \cite{piccinini2025model}, it remains unclear how these techniques compare to empirical controllers in \textit{closed-loop} operation, where feedforward controllers interact with the high-level \ac{mpc}, the low-level feedback loop, and the actual vehicle dynamics at the limits.

    \subsection{Related Work}
    Existing feedforward steering controllers for automated vehicles fall into two broad categories: analytical model-based and learning-based methods.

    Analytical approaches rely on vehicle dynamics theory \cite{Abe2009}. The kinematic single-track model provides a simple relationship between the steering angle and the target path curvature, with the vehicle's wheelbase as the only parameter \cite{Wischnewski2021}. Assuming linear tire slip and force, this model can be extended to a dynamic single-track model, where the understeer gradient captures steering behavior dependent on axle cornering stiffness and weight distribution \cite{Frendo2006}.
    For vehicles with nonlinear tire characteristics and locked differentials, the understeer gradient fails to capture the relationship between lateral acceleration and steering, described by the \ac{hd} \cite{Frendo2006}. This motivates the use of dynamic single- and double-track models with nonlinear tire formulations in \ac{mpc}, treating the steering angle as a control input \cite{Raji2022,Vazquez2020,Jahncke_2025_diff_MPC,Listov2023}. However, parameter identification for such models is complex and affected by uncertainty \cite{Voser2010,Corradini2025}.

    Learning-based methods learn the vehicle dynamics directly from data, typically without physical assumptions. This includes learning the inverse vehicle dynamics with feedforward \cite{Spielberg2019,Spielberg2022,Devineau2018}, \acp{cnn} and recurrent networks \cite{piccinini2023predictive}. 
    Neural vehicle models have also been embedded in \ac{mpc}, either by replacing the analytical dynamics \cite{Xiao2025,gao2024integrated,Rokonuzzaman2021} or by learning residual dynamics on top of physics-based models \cite{dikici2025learning,miao2025residual}. However, poor generalization beyond the training domain remains a key limitation, due to the lack of physical structure in general-purpose \acp{nn} \cite{miao2025residual,piccinini2025model,dikici2025learning}.

    Hybrid physics-guided learning approaches aim to bridge the gap between analytical and learning-based methods by embedding physical insights into \ac{nn} architectures. 
    \acp{pinn} use physics-based loss functions to enforce dynamic system equations \cite{Raissi2019,Meng2025,Zeipel2024}, while \acp{msnn} use domain knowledge to design the internal architecture \cite{piccinini2025road,da2020mental,Piccinini2023_physics_driven,piccinini2025model}. 
    These approaches have shown better generalization and sample efficiency than generic \acp{nn} \cite{Faroughi2024,piccinini2025model}. 

    To the best of our knowledge, no prior work provides a consistent closed-loop comparison of learning-based and analytical feedforward steering methods for autonomous racing at the handling limits.

    \subsection{Contributions}
    The main contributions of this work are:
    \begin{itemize}
        \item A benchmark of analytical (including our empirical \ac{hd}) and learning-based feedforward steering controllers for autonomous racing, within a validated hierarchical software stack.
        \item A new feedforward steering controller based on an empirical \ac{hd} formulation, to capture the velocity-dependent nonlinear steering behavior with minimal parametrization.
        \item A comparison of open- and closed-loop performance, showing that our \ac{ehd} approach performs well and that higher inverse dynamics prediction accuracy does not necessarily yield better closed-loop tracking performance or lap times.
    \end{itemize}

\section{Methodology}
This section provides an overview of the wider control framework in which the feedforward approaches are integrated, as well as a description of all benchmarked feedforward algorithms. The empirical approaches include a baseline and the proposed \ac{ehd} approach. The learning-based controllers include an \ac{msnn} and an \ac{lstm}.    
\subsection{Background and Control Framework}
\begin{figure}
    \centering
    \def\svgwidth{\columnwidth}
    {\footnotesize
    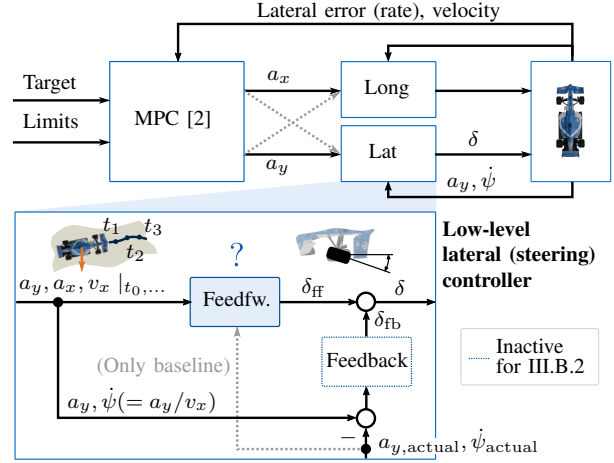}
    \caption{Overview of the hierarchical control architecture with the high-level \ac{mpc} and low-level feedforward-feedback steering controller.}
    \label{fig:hierarchical_control_algorithm}
\end{figure}

The tested feedforward approaches are integrated into 
\ifnum \isAnonymous=1
    a software stack for autonomous racing developed for competitions with full-scale cars.
\else
    the \ac{tam} software stack, which we used in the real-world 2025 Abu Dhabi Autonomous Racing League with full-scale cars \cite{hoffmann2026HeadtoHeadautonomousracing}.
\fi
 The hierarchical control module shown in Fig. \ref{fig:hierarchical_control_algorithm} consists of a high-level \ac{mpc} and low-level controllers. The \ac{mpc} computes longitudinal and lateral acceleration profiles to track a reference path and velocity provided by an upstream trajectory planner \cite{Wischnewski2021}. The \ac{mpc} acceleration output is executed by decoupled low-level lateral and longitudinal controllers \cite{Pitschi2025}. The lateral controller has a feedforward-feedback structure: the feedforward term aims to reduce feedback corrections by modeling the vehicle's inverse lateral dynamics, predicting the steering angle required to achieve the \ac{mpc} acceleration references. This section presents two empirical and two learning-based methods for this feedforward task.
\subsection{Baseline Empirical Controller}

The baseline feedforward controller employs an empirical strategy that incorporates multiple linear terms and has been extensively validated 
\ifnum \isAnonymous=1
    in real-world applications.
\else
    in real-world applications as part of the \ac{tam} team's software stack \cite{hoffmann2026HeadtoHeadautonomousracing}.
\fi
The feedforward steering angle $\delta_{\text{ff}}$ is determined from the lateral acceleration target $a_{y}$ as a sum of the kinematic steering angle $\delta_{\text{ack}} = \frac{a_{y}}{v_x^2} l$ with the vehicle's longitudinal velocity $v_x$ and wheelbase $l$, a term $\delta_{\text{ug}} = k_{\text{ug}} a_{y}$ representing a constant understeer gradient, and a component $\delta_{\text{long}} = k_{\text{long}} a_{x,\text{actual}} a_{y}$ that is proportional to the measured longitudinal acceleration $a_{x,\text{actual}}$ to account for longitudinal load transfer and combined tire force effects:

\begin{equation}
	\delta_{\text{ff}} = \delta_{\text{ack}} + \delta_{\text{ug}} + \delta_{\text{long}}.
\end{equation}

The feedforward components $\delta_{\text{ug}}$ and $\delta_{\text{long}}$ are further processed using a first-order low-pass filter. The weights $k_{\text{ug}}$ and $k_{\text{long}}$, as well as the low-pass time constants, can be adjusted to match the vehicle's dynamics or to achieve a specific control objective of the overall system. Additionally, the parameter $k_{\text{long}}$ is assigned different values for positive and negative longitudinal accelerations.

\subsection{Proposed \acl{ehd} Controller}
\begin{figure}[htbp]
    \centering
    \tiny
    \def\svgwidth{0.5\columnwidth}
    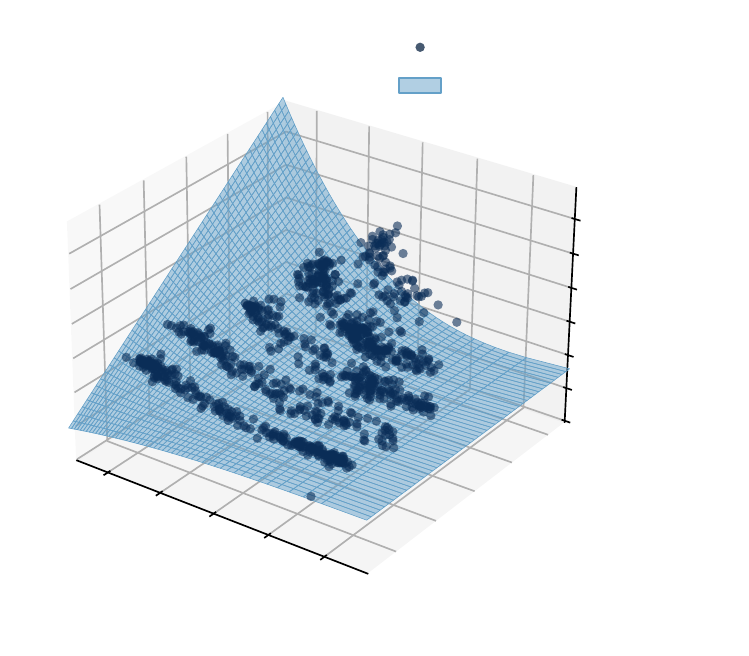%
    \includegraphics[width=0.5\columnwidth]{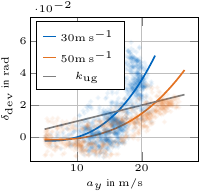}
    \caption{\textit{Left:} \ac{hd} with polynomial surface fit of data from simulated laps of the Abu Dhabi North track. Data is rounded to the nearest 10 $\si{\metre\per\second}$ for readability. \textit{Right:} Cross-sections of the polynomial surface fit compared to the static understeer gradient $k_\mathrm{ug}$ from the baseline.}
    \label{fig:handling_diagram}
\end{figure}

The \ac{hd} is an established representation of steady-state vehicle cornering behavior. It characterizes the relationship between lateral acceleration $a_y$ and the steering deviation $\delta_{\text{dev}}$, commonly associated with the understeer gradient metric $K=\frac{\partial \delta_{\text{dev}}}{\partial a_y}$. This deviation is defined as the difference between the measured and the kinematic steering angles. Mathematically:
\begin{equation}
    \delta_{\text{dev}} = \delta - \delta_{\text{ack}} = \alpha_{\text{f}} - \alpha_{\text{r}}
\end{equation}
with front/rear side slip angles $\alpha_{\text{f}/\text{r}}$. If the \ac{hd} slopes upward, the vehicle is understeering ($K>0$), if it slopes downward the vehicle is oversteering ($K<0$). The \ac{hd} for a single-track model with a linear tire model is a straight line. In practice, vehicles with nonlinear tires and locked differentials exhibit nonlinear \ac{hd}s that are dependent on the vehicle state. \cite{Frendo2006}\\
To capture this nonlinearity, we propose an \ac{ehd} approach based on a polynomial surface fit of the \ac{hd} over lateral acceleration $a_y$ and longitudinal velocity $v_x$ (see Fig. \ref{fig:handling_diagram}). The resulting polynomial function $\hat{\delta}_{\text{dev}}(a_y, v_x)$ then defines the feedforward steering command as
\begin{equation}
    \delta_{\text{ff}} = \delta_{\text{ack}} + \hat{\delta}_{\text{dev}} = \frac{a_yl}{v_x^2} + \hat{\delta}_{\text{dev}}.
    \label{eq:steering_angle_from_steer_dev_pred}
\end{equation}
The fitting function is given by the following polynomial, odd in $a_y$ and linear in $v_x$:
\begin{equation}
    \hat{\delta}_{\text{dev}}(a_y, v_x) = (k_{v1}v_x+k_{v0})(k_{a3}a_y^3 + k_{a1}a_y)
\end{equation}
with fitting parameters $k_{v1}, k_{v0}, k_{a3}, k_{a1}$.
For fitting, we transform this function as follows:
\begin{equation}
    z = \tilde{k}_{v1a3} xy + \tilde{k}_{a3} x + \tilde{k}_{v1a1} y + \tilde{k}_{a1},
\end{equation}
with transformed quantities $x = a_y^2$, $y = v_x$, $z = \frac{\hat{\delta}_{\text{dev}}}{a_y}$, and fitting parameters $\tilde{k}_{v1a3}=k_{v1}k_{a3}$, $\tilde{k}_{a3}=k_{a3}k_{v0}$, $\tilde{k}_{v1a1}=k_{v1}k_{a1}$, $\tilde{k}_{a1}=k_{a1}k_{v0}$. The quantity $z=\frac{\hat{\delta}_\text{dev}}{a_y}$ is the secant slope of the \ac{hd}, and thus describes the average understeer gradient for accelerations up to the current value.

\subsection{Model-Structured Neural Network Controller}
The \ac{msnn} computes the feedforward steering angle by embedding prior knowledge of the nonlinear vehicle dynamics directly into the network architecture. It approximates the quasi steady-state cornering behavior with local \ac{hd} models that activate in predefined regions of lateral acceleration, longitudinal acceleration, and velocity. This neural fuzzy logic is combined with fully connected layers that capture transient effects. Architectural details and derivations are provided in \cite{piccinini2025model}.

\subsection{
Long Short-Term Memory Network Controller}
Inspired by other \ac{lstm} feedforward implementations \cite{piccinini2023predictive}, this algorithm uses an \ac{lstm} network to approximate the inverse lateral dynamics of the vehicle. The \ac{lstm} allows for an efficient encoding of dynamic states along a time horizon. The network takes as input a horizon of $N=10$ velocities $v_x$ and accelerations $a_x,a_y$, interpolated from the \ac{mpc}-generated trajectory with a time interval of $0.01\,\mathrm{s}$. The sequence is processed by a single-layer \ac{lstm} to produce a 64-dimensional hidden state, which is passed through a fully connected layer to obtain $\hat{\delta}_\mathrm{dev}$. The feedforward steering command is then calculated by summing $\hat{\delta}_\mathrm{dev}$ and $\delta_\mathrm{ack}$ as shown in \eqref{eq:steering_angle_from_steer_dev_pred}. We use a dropout layer to randomly deactivate 10\% of the hidden state values at the output of the \ac{lstm} during each training iteration.
\begin{figure}[htbp]
    \centering
    \def\svgwidth{\linewidth}
    {
    \footnotesize
    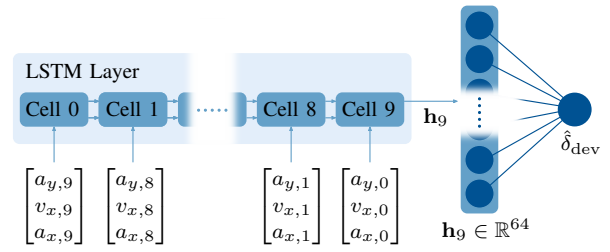
    }
    \caption{Schematic of the \ac{lstm} based feedforward control algorithm}
    \label{fig:lstm}
\end{figure}

\section{Results}

    We evaluate the four feedforward algorithms in open- and closed-loop settings. 
    In the open-loop evaluation, we compare the inverse dynamics modeling accuracy of the feedforward algorithms on a driving dataset. 
    The closed-loop evaluation assesses interaction with the remaining control module (see Fig.~\ref{fig:hierarchical_control_algorithm}) and system dynamics. 
    Data collection and closed-loop evaluation are conducted with a high-fidelity double-track vehicle model \cite{Sagmeister2024} of the Dallara EAV24 on the Abu Dhabi North circuit. 
    Model parameters are derived from real vehicle data and the simulation has been validated through extensive testing in the \ac{a2rl} and \ac{iac} competitions~\cite{hoffmann2026HeadtoHeadautonomousracing,Sagmeister2024}. 


    \subsection{Training and Open-Loop Performance}
    We collect the training data from a full-stack simulation of the vehicle with the baseline feedforward controller. The dataset consists of 26 laps, driven at race pace, with progressively increasing GG scale (i.e., scaling of longitudinal and lateral acceleration limits) as determined by the planning module \cite{Werner2025}. We utilize an 80/20 train-validation split for the \ac{ehd}, \ac{msnn}, and \ac{lstm} approaches.
	
\begin{figure}[t]
        \centering
        \includegraphics{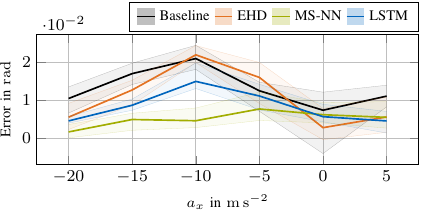}
        \vspace{2.5mm}
        {
        \footnotesize
        \hspace{1mm}
        \begin{tabular}{lccc}
            \toprule
            {Model} & {\ac{rmse}} & {\ac{mae}} & \ac{fvu} \\
            \midrule
            Baseline & 0.009176 & 0.006791 & 0.0591 \\
            EHD      & 0.007617 & 0.004452 & 0.0392\\
            MS-NN    & 0.005582 & 0.003514 & 0.0212\\
            LSTM     & 0.005525 & 0.003328 & 0.0206\\
            \bottomrule
        \end{tabular}
        }
        \caption{\textit{Top:} Steering direction normalized error, $\mathrm{sgn}(\delta)(\delta_\mathrm{ff}-\delta)$, across different longitudinal accelerations during cornering ($\rho\geq0.003\,\si{\radian\per\metre}$). 
		\textit{Bottom:} \ac{rmse}, \ac{mae} and \ac{fvu} of the steering angle prediction for the four feedforward approaches.}
        \label{fig:open_loop_statistics}
    \end{figure}

	The algorithms are evaluated on an unseen test set consisting of a single lap, with the \ac{rmse}, \ac{mae}, and \ac{fvu} of the predicted steering angle as metrics (see Fig.~\ref{fig:open_loop_statistics}). The proposed \ac{ehd} approach reduces \ac{rmse} by 17.0\% and \ac{mae} by 33.7\% over the baseline. The learning-based approaches show further improvements, with a 39.1\% decrease in \ac{rmse} and a 64.1\% decrease in \ac{fvu} for the \ac{msnn} relative to the baseline. The \ac{lstm} reaches the highest accuracy with an improvement of 39.7\% in \ac{rmse} and 65.1\% in \ac{fvu} over the baseline.
    
\begin{figure}
		\centering
		\includegraphics{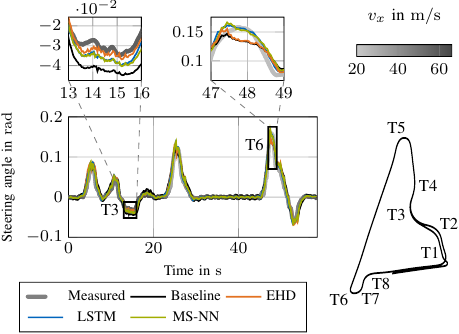}
		\caption{\textit{Left:} Open-loop steering angle prediction on the test lap with details showing predictions for a high-speed and a low-speed corner. \textit{Right:} Overview of the Abu Dhabi North track layout with corner numbers.}
        \label{fig:open_loop_timeseries}
	\end{figure}

    Figure \ref{fig:open_loop_timeseries} shows the steering predictions over the test lap. We focus on a high-speed turn (T3, $v_x\approx 55 \si{\metre\per\second}$) and a low-speed turn (T6, $v_x\approx 20 \si{\metre\per\second}$). 
	The baseline approach overestimates the steering angle in T3 by approximately 30\% and underestimates it in T6 by approximately 25\%. 
	The \ac{ehd} approach improves the performance in both corners, with the most accurate prediction across all approaches in T3 and a slight improvement on entry of T6 compared to the baseline. 
	This demonstrates the benefit of considering velocity dependence in our \ac{ehd}. 
	The learning-based methods are slightly less accurate compared to the \ac{ehd} in T3, but show significant improvements in T6. 
	Furthermore, we observe that the learning based approaches do not overestimate the steering angle during braking (negative $a_x$) to the extent that the baseline and \ac{ehd} approaches do (see Fig. \ref{fig:open_loop_statistics}).
	This can be attributed to the more complex model structure and additional features, such as longitudinal acceleration. Thus the learning-based approaches can capture more complex phenomena such as transient weight transfer and combined lateral-longitudinal tire slip dynamics.

\begin{figure}
        \includegraphics{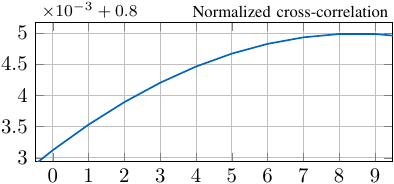}
		\includegraphics{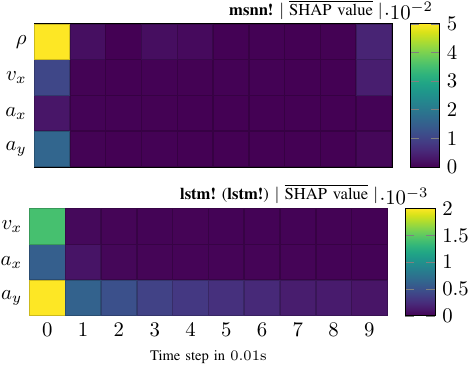}
		\caption{\textit{Top:} Normalized cross-correlation between lateral acceleration and steering angle. \textit{Middle and bottom:} Mean absolute SHAP importance scores for the features in the \ac{msnn} (middle) and the  \ac{lstm} (bottom) evaluated for 1000 samples. The peak in cross-correlation at 90 \si{\milli\second} appears to correlate with higher feature importance at step 9 of the \ac{msnn} \ac{shap} map. This is an indication of the \ac{msnn}'s ability to extract information of the transient effects in the system.}
        \label{fig:shap}
	\end{figure}

	By calculating the mean absolute \ac{shap} value for each input feature \cite{Lundberg2017}, we determine what information the learning-based approaches use to make their predictions. 
	Figure \ref{fig:shap} (middle and bottom) shows the mean absolute \ac{shap} values across 1000 random samples of the test dataset. 
	Both models rely most on the first \ac{mpc} step. 
	The current kinematic path curvature $\rho=\frac{a_y}{v_x^2}$ is the most important feature for the \ac{msnn}, followed by current lateral acceleration $a_y$, longitudinal velocity $v_x$ and longitudinal acceleration $a_x$. 
	With the exception of $\rho$, which is not provided as an input feature, the \ac{lstm} exhibits the same ranking of feature importance.

	Normalized cross-correlation between measured steering angle and lateral acceleration indicates that changes in steering angle precede changes in lateral acceleration by approximately 90 ms (see Fig. \ref{fig:shap} top).
	This delay is also reflected in the \ac{msnn}, which has a slight uptick in feature importance for the ninth time step in the input horizon (see Fig. \ref{fig:shap} middle).
	In contrast, the \ac{lstm} has a more balanced distribution of importance that does not peak at 90 ms (see Fig. \ref{fig:shap} bottom). 
	This suggests that the \ac{lstm} extracts its information from a broader time horizon, but does not seem to recognize the lag between the acceleration and steering angle to the extent as the \ac{msnn}. 
	This is due to the \ac{lstm} architecture's inherent bias towards inputs more closely connected to the hidden state output.

    \subsection{Closed-Loop Results}
	For closed-loop testing, all algorithms are tested on scenarios with increasing GG scale in the planner trajectory until failure. 
	We define failure with the lateral path tracking error threshold of $2.2\, \si{\metre}$. 
	For repeatability, we run all tests with the deterministic \ac{rslcpp} framework~\cite{sagmeister2026rslcppdeterministicsimulations}.

	\subsubsection{Evaluation with Feedback Control}
\begin{figure}
    	\centering
    	\includegraphics{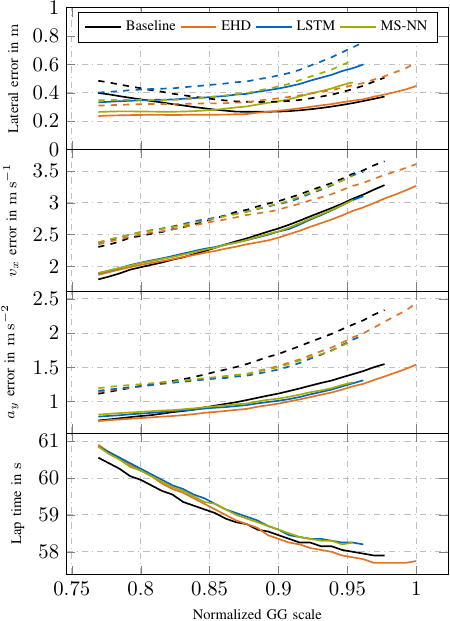}
    	\caption{Performance metrics for closed-loop evaluation at different GG values. The GG values indicate how the acceleration limits are scaled relative to the most demanding scenario that was still completed successfully. \textit{Dashed:} \ac{rmse} values; \textit{Solid:} \ac{mae} values.}
    	\label{fig:closed_loop_statistics_comparison}
    \end{figure}

    In the first series of tests we run the four feedforward architectures with the feedback active (see Fig. \ref{fig:hierarchical_control_algorithm}). All software parameters are kept constant with the exception of the feedforward architectures. 
	The proposed \ac{ehd} achieves the highest GG scale of all approaches, followed by the baseline, the \ac{lstm} and the \ac{msnn} (see Fig. \ref{fig:closed_loop_statistics_comparison}). 
	A similar ranking can be observed for overall lap time, with the \ac{ehd} achieving the fastest lap time of 57.7 \si{\second}, followed by the baseline with 57.9 \si{\second}. 
	The learning-based approaches both reach a lap time of 58.2 \si{\second}. 
	This is in spite of an improvement in lateral acceleration tracking compared to the baseline. 
	Lateral path tracking performance deteriorates for all approaches as the GG scale increases, with exception of the baseline approach, which shows a slight improvement up until 0.88 with subsequent deterioration to failure at a GG scale of 0.98 relative to the maximum achieved by the \ac{ehd}. 
    This sweet-spot reflects the manual tuning that is optimized for stability at high GG scale. 
	Velocity tracking is relatively consistent across all approaches with the exception of the \ac{ehd}, which slightly outperforms the other methods from GG 0.85 and above. 
    
\begin{figure}
    	\centering
    	\includegraphics{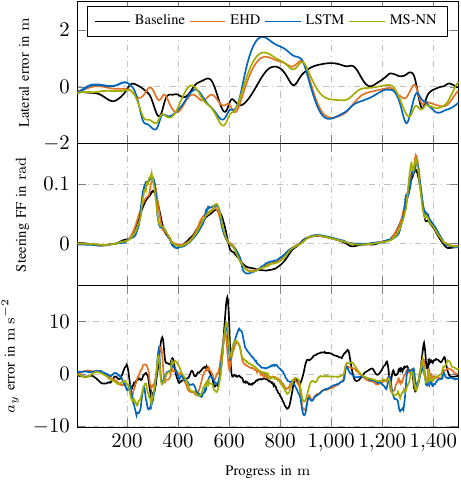}
    	\caption{Comparison of lateral error and steering angle for the track section before turn 1 until after turn 5 with feedforward and feedback control in closed-loop operation.}
    	\label{fig:closed_loop_timeseries_comparison}
    \end{figure}

	The observed effects can be better understood by looking at the time-series data of the four approaches in closed-loop operation (see Fig. \ref{fig:closed_loop_timeseries_comparison}).
	Both empirical approaches have the tendency to steer more aggressively early in the corner compared to the learning-based approaches. 
	This is especially pronounced in the baseline approach, which has small acceleration errors in the direction of the turn at 200 \si{\metre} (entry turn 1), 400 \si{\metre} (entry turn 2), and 1200 \si{\metre} (entry turn 5), compared to the other approaches. 
	The \ac{ehd} approach has a similar characteristic for low-speed corners (e.g., at 200 \si{\metre}), but shows a steering profile more comparable to the learning-based approaches in high-speed corners (e.g., at 600 \si{\metre}). 
	Even though the learning-based approaches use future acceleration targets for their predictions, they appear to have a more \textit{delayed} steering behavior. This is because the effect is not temporal, but rather due to the pattern observed in our open-loop tests: The learning-based approaches predict a lower steering angle during braking at corner entry compared to the empirical approaches (see Fig. \ref{fig:open_loop_statistics}).

	\subsubsection{Evaluation without Feedback Control} 
\begin{figure}
    	\centering
    	\includegraphics{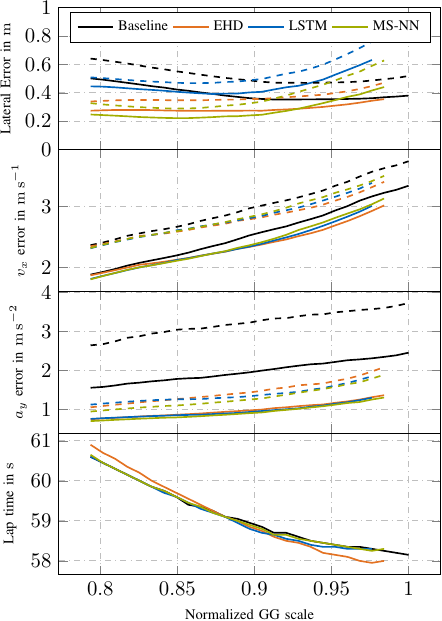}
    	\caption{\ac{rmse} and \ac{mae} of performance metrics for closed-loop evaluation without feedback control at different GG values. \textit{Dashed:} \ac{rmse} values; \textit{Solid:} \ac{mae} values.}
    	\label{fig:closed_loop_no_fb_statistics_comparison}
    \end{figure}

\begin{figure}
    	\centering
    	\includegraphics{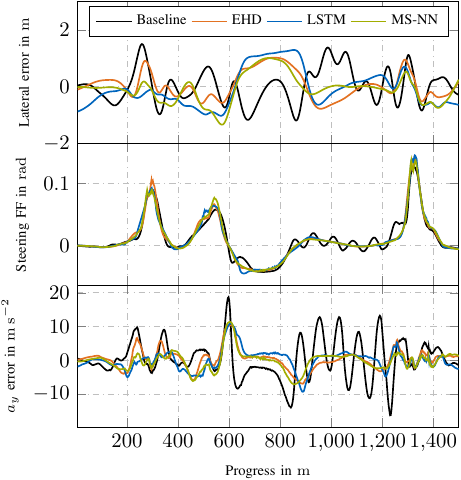}
    	\caption{Comparison of lateral error, steering angle and lateral acceleration for the track section before turn 1 until after turn 5 without feedback control in closed-loop operation.}
    	\label{fig:closed_loop_no_fb_timeseries_comparison}
    \end{figure}

To further isolate the effect of the feedforward algorithms, we deactivate the feedback steering control loop and repeat the closed-loop evaluation with only the high-level \ac{mpc} and the feedforward controller in the loop. 
	The ranking of the controllers is mostly consistent with the previous closed-loop evaluation, with the \ac{ehd} approach showing the best overall lap time, followed by the baseline, the \ac{lstm} and the \ac{msnn} (see Fig. \ref{fig:closed_loop_no_fb_statistics_comparison}). 
	The \ac{ehd} and the learning-based methods show significant improvements in lateral acceleration tracking compared to the baseline, which does not necessarily translate to better lateral path tracking or lap time. 
	As in the closed-loop testing with low-level feedback, the baseline controller again exhibits an aggressive turn-in behavior, with more pronounced lateral acceleration overshoots as the inner feedback control loop is deactivated and errors need to be corrected by the slower \ac{mpc} (see Fig.~\ref{fig:closed_loop_no_fb_timeseries_comparison}). In contrast, the learning-based approaches initiate the turn later as their learned inverse dynamics models predict steering more accurately at low lateral acceleration. We observe this as a relative lack of lateral acceleration overshoots on entry and exit of turn 3 (see Fig.~\ref{fig:closed_loop_no_fb_timeseries_comparison} bottom at 600 and 800 \si{\metre}). Consequently, the baseline approach takes the turns more to the inside (see Fig.~\ref{fig:closed_loop_no_fb_timeseries_comparison} top at 650 \si{\metre}), leaving space for drifting to the outside and also reducing the overall curvature as the car effectively cuts the corner. 
    The downside of this is that the baseline approach is prone to oscillations in high-speed turns, as small acceleration targets are translated into greater steering corrections compared to the velocity-dependent progressive steering models of the \ac{ehd} (see Fig.~\ref{fig:handling_diagram}) and learning-based approaches. As a result, we not only observe large oscillations in the time-series but also quantify them with an elevated root mean square lateral jerk of 17.28 \si{\metre\per\second\cubed} compared to 10.54 \si{\metre\per\second\cubed} in the \ac{ehd}, 10.42 \si{\metre\per\second\cubed} in the \ac{lstm}, and 10.14 \si{\metre\per\second\cubed} in the \ac{msnn}. 

	\subsubsection{Iterative Fine-Tuning} 
\begin{figure}
    	\centering
    	\includegraphics{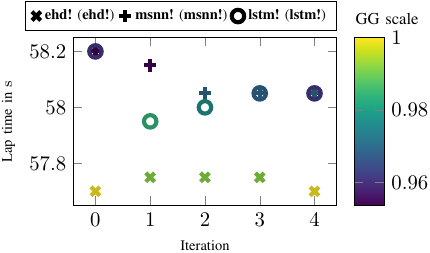}
    	\caption{Lap time evolution under iterative fine-tuning (retraining after closed-loop deployment) of the \ac{ehd}, \ac{msnn}, and \ac{lstm} methods. Although \ac{msnn} and \ac{lstm} improve with retraining, our \ac{ehd} achieves the lowest closed-loop lap times.}\label{fig:model_refinement}
    \end{figure}

In the final set of tests we simulate a use case where the feedforward model is iteratively fine-tuned based on data collected over multiple runs in closed-loop operation during testing. 
	After each run, we initialize the learning-based approaches with the saved parameters and resume training with reduced learning rate on the newly collected data. 
	For the \ac{ehd} approach, we extend the dataset with the newly collected data and refit the polynomial surface. 
	Both learning-based approaches seem to benefit from iterative fine-tuning, with a lap time improvement of 0.15 $\si{\second}$ for both approaches after 4 iterations of deployment and training (see Fig. \ref{fig:model_refinement}). 
	The \ac{msnn} shows a more consistent improvement across iterations, while the \ac{lstm} shows a significant gain in the first iteration and then deteriorates and plateaus to the same lap time as the \ac{msnn}. 
	The \ac{ehd} approach does not show a clear trend: its structure is not as flexible as the learning-based approaches, which makes it less prone to overfitting, but also limits its ability to learn from additional data. 
	Despite their fine-tuning, the learning-based controllers do not match the \ac{ehd} approach in closed-loop operation.
	This suggests that an inverse vehicle dynamics model to track the lateral acceleration targets from the \ac{mpc} may not be sufficient to achieve optimal performance in terms of lap time and path tracking.
	The simplified kinematic assumptions used in the trajectory planning and \ac{mpc} modules cannot accurately capture the transient vehicle dynamics but only the acceleration limits (GGGV diagram and GG scale). Thus, the planned acceleration profiles may not perfectly match the desired path. In contrast to the baseline and \ac{ehd} relying on instantaneous targets, the \ac{lstm} and \ac{msnn} with their trajectory target input features might be more sensitive to this mismatch between kinematic assumptions and transient dynamics of the vehicle.   
\section{Conclusion and Outlook}
\label{sec:conclusion}
We compare two empirical and two learning-based feedforward steering control approaches.

In open-loop evaluation, learning-based approaches outperform the empirical approaches in terms of prediction accuracy. The \ac{ehd} approach does not match the performance of the learning-based approaches, but outperforms the baseline. Both the \ac{msnn} and the \ac{lstm} appear to capture inverse dynamics more precisely under braking conditions, likely due to their ability to learn load transfer and combined slip effects from data.

The conclusions drawn from the closed-loop evaluation are more nuanced. The \ac{ehd}, the \ac{msnn}, and the \ac{lstm} all lower lateral acceleration errors relative to the baseline. However, this does not translate to improved tracking error or lap time for all approaches. 
The \ac{ehd} approach improves lap time and robustness, while the \ac{msnn} and \ac{lstm} do not. 

We attribute this phenomenon to the following mechanism: the empirical approaches overestimate the steering angle during braking. In closed-loop, this translates to earlier turn initiation, effectively cutting corners and reducing the driven path curvature. This results in greater stability (w.r.t. target trajectory) and faster lap times, despite higher open-loop prediction errors.

The findings highlight the importance of closed-loop evaluation and suggest that focusing on acceleration may not align with global performance objectives. 
The planner and the \ac{mpc} generate acceleration targets with simplified kinematic assumptions and scaled steady-state acceleration limits. 
This introduces a mismatch between the acceleration targets and the actual transient vehicle dynamics. In this setting, the structured and low-variance nature of the \ac{ehd} approach can be advantageous compared to more flexible learning-based models.

Incorporating direct path or yaw rate information, instead of acceleration-derived kinematic curvature, into the learning-based approaches could allow for steering behavior that better aligns with overall tracking objectives. 
Another potential strategy is to use adaptive or online learning methods that account for controller interaction and optimize performance directly in closed loop.
    
	
\ifnum \isAnonymous=0
	\section*{Authors' Contributions}
	Georg Jank: Conceptualization, Methodology, Software, Visualization, Writing – original draft \& review \& editing. 
	Mattia Piccinini: Conceptualization, Methodology, Software, Writing - original draft \& review \& editing. 
	Sebastian Wenk: Conceptualization, Methodology, Investigation. 
	Phillip Pitschi: Visualization, Investigation, Methodology, Writing - original draft \& review \& editing.
	Johannes Betz \& Boris Lohmann: Supervision, Writing – review \& editing, Funding acquisition, Project administration.
\fi

	\bibliographystyle{IEEEtran}
	\bibliography{stylecustomization, root} 
	
\end{document}